\documentclass[10pt,twocolumn,letterpaper]{article}

 \usepackage{iccv}              % To produce the CAMERA-READY version
\usepackage{xcolor}
\usepackage{tabularx}

\usepackage{array}  % Needed for custom column types
\newcolumntype{C}[1]{>{\centering\arraybackslash}m{#1}}  % Centered horizontally and vertically

\usepackage{graphicx} % for \resizebox
\usepackage{array}    % for better table formatting

\definecolor{iccvblue}{rgb}{0.21,0.49,0.74}
\usepackage[pagebackref,breaklinks,colorlinks,allcolors=iccvblue]{hyperref}

\def\paperID{*****} % *** Enter the Paper ID here
\def\confName{ICCV}
\def\confYear{2025}

\title{VelocityNet: Real-Time Crowd Anomaly Detection via Person-Specific Velocity Analysis}
\author{
Fatima AlGhamdi \and
Omar Alharbi \and
Abdullah Aldwyish \and
Raied Aljadaany \and
Muhammad Kamran J Khan \and
Huda Alamri \and
Saudi Data and Artificial Intelligence Authority (SDAIA)\\
{\tt\small falghamdi, oalharbi, aaldwyish, raljadaany, mkkhan, haamri@ncai.gov.sa}
}

\nocite{*}
\begin{document}
\maketitle
\begin{abstract}

Detecting anomalies in crowded scenes is challenging due to severe inter-person occlusions and highly dynamic, context-dependent motion patterns. Existing approaches often struggle to adapt to varying crowd densities and lack interpretable anomaly indicators. To address these limitations, we introduce VelocityNet, a dual-pipeline framework that combines head detection and dense optical flow to extract person-specific velocities. Hierarchical clustering categorizes these velocities into semantic motion classes (halt, slow, normal, and fast), and a percentile-based anomaly scoring system measures deviations from learned normal patterns. Experiments demonstrate the effectiveness of our framework in real-time detection of diverse anomalous motion patterns within densely crowded environments.

%Detecting anomalies in pedestrian dynamics is a challenging computer vision problem, especially in densely crowded scenes, due to severe interperson occlusions and stringent real-time processing constraints. Existing approaches often fail to adapt to dynamically varying crowd densities and typically lack interpretable anomaly indicators, significantly limiting their practical effectiveness.
%To address this challenge, we introduce VelocityNet, a crowd anomaly detection framework that employs a dual-pipeline: video processing architecture integrating head detection and dense optical flow estimation to generate person-specific regions of interest. VelocityNet utilizes hierarchical clustering and percentile-based anomaly scoring, which effectively identify motion anomalies across varying densities. Experimental results demonstrate the framework's effectiveness in distinguishing diverse anomalous motion patterns in dense crowds.
\end{abstract}

\section{Introduction}

Anomaly detection is a fundamental task in computer vision, aiming to identify events or behaviors that deviate from established patterns without extensive supervision. Early anomaly detection methods relied on statistical models such as Gaussian mixture models (GMMs)~\cite{stauffer1999adaptive} and traditional feature extraction techniques like optical flow or Histograms of Oriented Gradients (HOG), typically combined with classifiers such as Support Vector Machines (SVMs)~\cite{adam2008robust}. While these methods provided initial success, their performance significantly declined in complex real-world environments characterized by variability, occlusions, and dynamic behaviors.

Recent advances in deep learning have substantially improved anomaly detection capabilities. Approaches using Autoencoders~\cite{hasan2016learning} and Generative Adversarial Networks (GANs)~\cite{ravanbakhsh2017abnormal} have emerged, detecting anomalies through reconstruction errors or inconsistencies in predicted video frames~\cite{liu2018future}. These methods benefit from data-driven learning that better captures the complexity and variability inherent in real-world scenes.

Among various anomaly detection domains, densely crowded environments represent a uniquely challenging yet critical scenario. Detecting anomalies within high-density crowds is difficult due to two primary factors: (1) severe occlusions, which obscure individual appearances and complicate tracking, and (2) highly dynamic, context-dependent motion patterns, where the definition of "normal" motion can vary dramatically depending on crowd density and spatial context.

Despite extensive research in anomaly detection~\cite{sultani2018real,vu2019robust,nguyen2019anomaly,georgescu2021background}, crowded scenes remain under-addressed, primarily due to lack of suitable datasets and models optimized for dense scenarios. Existing datasets are often limited in crowd density, diversity, and annotation detail, impeding progress in training robust and generalizable models. Moreover, stringent real-time constraints in practical deployment environments restrict model complexity, requiring solutions to be both computationally efficient and highly accurate.

In this paper, we propose a novel framework specifically designed to address anomaly detection in dense crowd scenarios. Our approach leverages head detection and dense optical flow estimation to analyze crowd motion at an individual level, categorizing motion patterns into semantically interpretable groups (halt, slow, normal, fast). We introduce an adaptive velocity-based anomaly scoring mechanism that automatically adjusts to varying crowd densities, allowing for context-sensitive anomaly identification. The proposed system achieves real-time performance, effectively overcoming previous limitations, and provides interpretable outputs suited for practical deployments.

Our main contributions are summarized as follows:

\begin{itemize}
    \item A dual-pipeline architecture combining head detection and dense optical flow for person-specific velocity estimation.
    \item Hierarchical clustering of velocities into semantic motion categories (halt, slow, normal, fast) for interpretable anomaly detection.
    \item A density-aware, percentile-based anomaly scoring mechanism for real-time anomaly detection in crowded scenes.
\end{itemize}

\section{Related Work}

Our work relates primarily to several research areas: Anomaly Detection in Videos, Velocity Estimation in Crowded Scenes, and Motion Representation and Analysis. Below, we review each of the research areas.

\subsection{Anomaly Detection in Videos}

Recent advances in video anomaly detection predominantly utilize deep-learning-based methods, including Autoencoders \cite{hasan2016learning}, Generative Adversarial Networks (GANs) \cite{ravanbakhsh2017abnormal}, and transformer architectures \cite{huang2022motion}. These methods typically detect anomalies through reconstruction errors or inconsistencies in predicted video frames or motion fields \cite{liu2018future,huang2022motion}. Benchmarks such as CUHK Avenue~\cite{lu2013abnormal}, ShanghaiTech~\cite{luo2017revisit}, and UCSD Ped2~\cite{wang2010anomaly} are commonly used to evaluate anomaly detection methods. However, these datasets mostly feature moderate crowd densities and clear anomalies such as unexpected actions or intrusions.
Furthermore, self-supervised multi-task approaches, such as SSMTL++, which incorporates an updated backbone and enhanced proxy tasks, consistently achieve state-of-the-art results on Avenue, ShanghaiTech, and UBnormal, thereby highlighting the significance of multi-task supervision in VAD\cite{barbalau2023ssmtlpp}. Our work specifically addresses anomaly detection in highly challenging dense crowd scenarios characterized by significant occlusions and subtle abnormal motions.

\subsection{Velocity Estimation in Crowded Scenes}

Velocity has been a crucial indicator of anomalous behavior, especially in crowd analysis. Early approaches utilized velocity-based features derived from optical flow to detect abnormal movements such as unusually rapid or halted pedestrians \cite{mehran2009abnormal,mahadevan2010anomaly}. More recent methods explicitly incorporate velocity and pose attributes for improved anomaly detection accuracy \cite{reiss2022anomaly}.
Despite the strong performance of velocity cues, prior work often ignored context-aware categorization and density-aware anomaly definitions. We address this by hierarchically clustering velocities into interpretable groups (halt, slow, normal, fast) and defining anomalies relative to local density. Closest to our approach, Reiss \& Hoshen combine velocity and pose with density-based scoring, achieving SOTA on Ped2, Avenue, and ShanghaiTech and reinforcing the interpretability of velocity-centric cues \cite{reiss2025attributevad}.

\subsection{Motion Representation and Analysis}

Understanding complex scene dynamics relies on motion representation. Many methods use optical flow or predicted motion to detect anomalies via frame prediction errors or inconsistencies \cite{liu2018future,huang2022motion}. Transformer-based approaches\cite{huang2022motion}, while effective, use deep sequence modeling, hindering interpretability and increasing computational complexity. Our method uses direct optical flow analysis with simple clustering for clear, interpretable motion descriptions and real-time anomaly detection. Additionally, SpeedNet learns a self-supervised "speediness" representation, proving speed is a meaningful, learnable attribute \cite{benaim2020speednet}.

\section{Methodology}

We propose \textbf{VelocityNet}, a crowd anomaly detection framework designed to identify velocity-based anomalies in densely crowded scenes. Given live video input, it outputs interpretable per-person motion categories and anomaly scores.

%We propose \textbf{VelocityNet}, a crowd anomaly detection framework designed to identify velocity-based anomalies in densely crowded scenes. Given input video frames, it outputs interpretable motion categories and anomaly scores at the individual level.

\begin{figure*}
    \centering
    \includegraphics[width=1\linewidth]{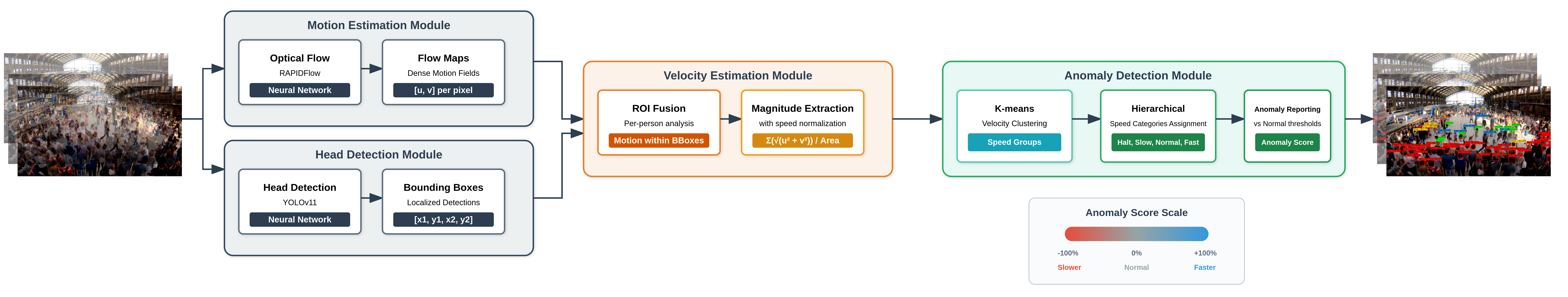}
    % \caption{Architecture overview of VelocityNet: Two parallel streams—head detection and dense optical flow—process the input simultaneously. These outputs merge to produce per-person velocity descriptors, which are then clustered and scored to detect anomalies in real time.}

    \caption{Architecture overview of VelocityNet: Two parallel streams—head detection and dense optical flow—process the input simultaneously. These outputs merge to produce per-person velocity descriptors, which are clustered during training to establish normal behavior boundaries. During inference, computed anomaly scores are compared against these predefined normal boundaries for real-time anomaly detection.}
   \label{fig:arch}
\end{figure*}

%-------------------------------------------------------------------------
\subsection{Overview}

Figure~\ref{fig:arch} presents \textit{VelocityNet}, which processes live video through two parallel streams, then merges results for anomaly analysis. First, the \textit{Motion Estimation Module} computes dense optical flow between incoming frames, capturing pixel-level motion across the scene. Simultaneously, the \textit{Head Detection Module} operates on raw frames to detect and localize heads, even under heavy occlusion, providing individual Regions of Interest (ROIs). These streams converge in the \textit{Velocity Estimation Module}, where flow is cropped to each head ROI, averaged to estimate raw per-person velocity, and normalized to account for perspective. Finally, the \textit{Anomaly Detection Module} clusters normalized velocities, applies density-aware adjustments, and assigns percentile-based anomaly scores.

In the following sections, we discuss each module in detail.

\subsection{Motion Estimation Module}

For temporal motion modeling of detected objects, we employ dense optical flow estimation to compute pixel-wise displacement vectors between consecutive frame pairs $(I_{t-1}, I_t)$. Our approach leverages RAPIDFlow~\cite{morimitsu2024rapidflow}, a recurrent all-pairs field transforms architecture that integrates NeXt1D convolution blocks within a fully recurrent pyramid structure to achieve computational efficiency while maintaining high estimation fidelity.

The preprocessing pipeline standardizes input frames to $1280\times720$ resolution, followed by pixel intensity normalization to the unit interval $[0, 1]$. The network outputs a dense flow field $\mathbf{F} \in \mathbb{R}^{H \times W \times 2}$, where each spatial location $(x, y)$ corresponds to a 2D displacement vector:

\begin{equation}
\mathbf{F}(x, y) = \begin{bmatrix} u(x, y) \\ v(x, y) 
\end{bmatrix}
\label{eq:motion-vector}
\end{equation}
representing the horizontal and vertical motion components, respectively. This dense correspondence field enables robust tracking of object  dynamics across temporal sequences while maintaining real-time processing capabilities through the recurrent pyramid architecture's computational optimizations.
%-------------------------------------------------------------------------

\subsection{Head Detection Module}

%To detect individuals in dense crowds effectively, VelocityNet focuses on a head-centric detection strategy. Heads are more consistently visible than full bodies in crowded environments, making them more reliable for tracking even under heavy occlusion.
   
%To focus exclusively on human motion analysis while mitigating common detection challenges such as inter-person occlusion and partial body visibility, 

To detect individuals in dense crowds effectively, VelocityNet focuses on a head-centric detection strategy. Heads are more consistently visible than full bodies in crowded environments, making them more reliable for tracking even under heavy occlusion. We employ this using \textbf{YOLO11} object detection architecture~\cite{yolo11_ultralytics}. Head regions exhibit superior detectability compared to full-body bounding boxes due to their reduced susceptibility to occlusion events and consistent appearance across varying poses and viewpoints.

For each detected head instance $j$ in frame $I_t$, we extract the bounding box coordinates:
\begin{equation}
\mathbf{B}_j^{(t)} = \{x_{\text{min}}, y_{\text{min}}, x_{\text{max}}, y_{\text{max}}\}_j^{(t)}
\label{eq:bbox-coordinate}
\end{equation}
where $(x_{\text{min}}, y_{\text{min}})$ and $(x_{\text{max}}, y_{\text{max}})$ define the top-left and bottom-right corners of the axis-aligned bounding rectangle, respectively. These spatial coordinates serve as regions of interest (ROIs) for subsequent motion analysis and temporal correspondence establishment.

%-------------------------------------------------------------------------

%\subsection{Pre-processing}

\subsection{Velocity Estimation Module}

In this module, we convert pixel-level motion into per-person velocity descriptors and correct for perspective distortion.

% For each detected head \(j\) in frame \(t\), the flow field from the Motion Estimation Module is cropped according to the bounding box \(\mathbf{B}_j^{(t)}\) to form a focused motion subregion. Within this ROI, each pixel’s displacement vector is denoted \(\mathbf{f}_{i,j}^{(t)}\).

% We compute the raw average motion magnitude for person \(j\):

% \begin{equation}
% \bar{m}_j^{(t)} = \frac{1}{|\mathbf{B}_j^{(t)}|} \sum_{(x, y) \in \mathbf{B}_j^{(t)}} \| \mathbf{f}_{i,j}^{(t)} \|_2,
% \end{equation}

% where \(\|\cdot\|_2\) is the Euclidean norm, and \(|\mathbf{B}_j^{(t)}|\) denotes the number of pixels in the bounding box.

Following dense optical flow estimation between consecutive frame pairs $(I_{t-1}, I_t)$, we perform spatial cropping of the flow field using detected head bounding boxes to isolate human-centric motion regions. This ROI-based extraction eliminates extraneous background motion and focuses computational resources on subjects of interest.

For each detected person instance $j$ with bounding box $\mathbf{B}_j^{(t)}$, we extract the corresponding flow subregion $\mathbf{F}_j^{(t)} \subset \mathbf{F}^{(t)}$ and compute the per-pixel motion magnitude:

\begin{equation}
m_{i,j}^{(t)} = \|\mathbf{f}_{i,j}^{(t)}\|_2 = \sqrt{(u_{i,j}^{(t)})^2 + (v_{i,j}^{(t)})^2}
\label{eq:pixel-magnitude}
\end{equation}

where $\mathbf{f}_{i,j}^{(t)} = (u_{i,j}^{(t)}, v_{i,j}^{(t)})$ represents the displacement vector at pixel location $i$ within person $j$'s bounding box at frame $t$.

To obtain a representative motion descriptor for each person, we compute the spatial average of magnitudes across all pixels within the bounding box:

\begin{equation}
\bar{m}_j^{(t)} = \frac{1}{|\mathbf{B}_j^{(t)}|} \sum_{i \in \mathbf{B}_j^{(t)}} m_{i,j}^{(t)}
\label{eq:avg-magnitude}
\end{equation}

where $|\mathbf{B}_j^{(t)}|$ denotes the cardinality of pixels within the bounding box region.

Next, to produce depth-invariant velocities, we apply one of two normalization techniques:

\textbf{Area-based normalization}
\begin{equation}
m_{\mathrm{norm}, j}^{(t)} = \frac{\bar{m}_j^{(t)}}{|\mathbf{B}_j^{(t)}|}
\end{equation}

\textbf{Unified-scale normalization}  
Using a predetermined target box size \(p\), we calculate a scale factor:
% \begin{equation}
% d_j^{(t)} = \sqrt{\frac{|\mathbf{B}_j^{(t)}|}{p^2}}
% \end{equation}

\begin{equation}
s_j^{(t)} = \frac{p^2}{|\mathbf{B}_j^{(t)}|}
\label{eq:area-scale}
\end{equation}

After resampling the cropped motion magnitude map to \(p \times p\) using bilinear interpolation, we apply scale-aware intensity adjustment to preserve motion consistency:

\begin{equation}
m_{\text{adj}, j}^{(t)} = m_{\text{rescaled}, j}^{(t)} \cdot 
\begin{cases}
s_j^{(t)}, & \text{if } s_j^{(t)} > 1 \\
1 / s_j^{(t)}, & \text{otherwise}
\end{cases}
\label{eq:scaling-adjustment}
\end{equation}

The final normalized motion descriptor is the spatial average over the adjusted patch:

\begin{equation}
m_{\text{norm}, j}^{(t)} = \frac{1}{p^2} \sum_{i=1}^{p^2} m_{\text{adj}, j,i}^{(t)}
\label{eq:normalized-magnitude}
\end{equation}

% The normalization process applies scale-adaptive resampling to achieve target dimensions $p \times p$, using anti-aliasing followed by decimation for downsampling ($d_j > 1$) with pixel intensities scaled by $1/d_j$, 
% \begin{equation}
% m_{\mathrm{norm}, j}^{(t)} = \frac{\bar{m}_j^{(t)}}{d_j^{(t)}}
% \end{equation}
% and bilinear interpolation with intensity scaling by factor $d_j$ for upsampling ($d_j < 1$).

% \begin{equation}
% m_{\mathrm{norm}, j}^{(t)} = \bar{m}_j^{(t)}{d_j^{(t)}}
% \end{equation}

The resulting normalized velocity \(m_{\mathrm{norm}, j}^{(t)}\) is a depth-invariant descriptor for each individual, passed onward to the next module.

\subsection{Anomaly Detection Module}

The anomaly detection module in \textbf{VelocityNet} consists of multiple interconnected components designed to categorize pedestrian motion and identify deviations from normal behavior. This is achieved through unsupervised clustering, semantic grouping, density-aware modeling, and interpretable anomaly scoring. Below, we describe each component in detail.

\subsubsection{Unsupervised Motion Clustering}
To identify recurring motion patterns, we first aggregate normalized motion magnitudes from all detected individuals across the video sequences into a unified feature vector \(\mathbf{M} = \{m_1, m_2, \ldots, m_N\}\), where \(N\) is the total number of motion observations. To ensure temporal continuity in multi-scene datasets, we exclude transitional frames at the boundaries between scenes to avoid spurious motion artifacts.

We employ K-means clustering to group the motion descriptors and use the elbow method to determine the optimal number of clusters \(k\), based on minimizing within-cluster sum of squares (WCSS):
\[
\text{WCSS}(k) = \sum_{i=1}^{k} \sum_{\mathbf{m} \in C_i} \|\mathbf{m} - \boldsymbol{\mu}_i\|^2
\]
where \(C_i\) is the \(i\)-th cluster and \(\boldsymbol{\mu}_i\) its centroid. The optimal \(k\) is selected by identifying the inflection point of the WCSS curve:
\[
k = \arg\max_{k} \left| \frac{d^2\text{WCSS}(k)}{dk^2} \right|
\]

% We note that silhouette analysis consistently favors fewer clusters (2--3), but this under-segments motion patterns, reducing the resolution needed for fine-grained behavioral analysis.

While silhouette coefficient analysis was considered as an alternative metric, it consistently favored lower cluster counts (2-3 clusters) compared to the elbow method (7-8 clusters), thereby limiting the granularity of motion pattern discrimination essential for fine-grained behavioral analysis.

\subsubsection{Semantic Grouping via Hierarchical Clustering}
To map K-means clusters to interpretable motion categories, we perform hierarchical agglomerative clustering based on cluster-level statistics. Each K-means cluster \( C_i \) is represented by a motion descriptor vector:

\begin{equation}
\boldsymbol{\phi}_i = \begin{bmatrix}
\mu_i \\
\sigma_i
\end{bmatrix}
\label{eq:cluster-features}
\end{equation}

where \( \mu_i \) and \( \sigma_i \) denote the mean and standard deviation of the motion magnitudes within cluster \( C_i \), respectively.

To merge similar clusters, we employ Ward's linkage criterion, which minimizes the total within-cluster variance. The pairwise distance between clusters \( C_i \) and \( C_j \) is defined as:

\begin{equation}
d(C_i, C_j) = \sqrt{ \frac{ |C_i| \cdot |C_j| }{ |C_i| + |C_j| } } \cdot \left\| \boldsymbol{\phi}_i - \boldsymbol{\phi}_j \right\|_2
\label{eq:ward-linkage}
\end{equation}

where \( |C_i| \) and \( |C_j| \) represent the number of motion vectors (or pixel samples) in each cluster, and \( \| \cdot \|_2 \) denotes the Euclidean norm between cluster descriptors.

% This process yields four semantic motion categories: \textbf{halt}, \textbf{slow}, \textbf{normal}, and \textbf{fast}, ordered by ascending velocity. Reliable separation of the \textbf{normal} category is crucial, as it serves as the reference baseline for anomaly scoring.

This process yields four semantic motion categories \textbf{halt}, \textbf{slow}, \textbf{normal}, and \textbf{fast}, arranged in ascending velocity order. The algorithm adaptively determines group membership without predetermined cluster count constraints, though performance degrades with insufficient K-means granularity ($k < 3$), as this prevents adequate representation of the \textbf{normal} velocity baseline required for anomaly threshold establishment.

\subsubsection{Density-Aware Modeling}
Crowd density significantly impacts expected pedestrian velocity. In low- to medium-density scenes, individuals typically walk at consistent, unconstrained speeds. In contrast, high-density environments exhibit reduced motion due to physical restrictions and visual occlusions.

To account for this variation, we categorize input scenes into two regimes:
\begin{itemize}
    \item Low-to-medium density
    \item High density
\end{itemize}

Each regime is assigned a dedicated model trained only on its respective data subset. This specialization improves accuracy and robustness, ensuring that slow-but-normal motion in high-density contexts is not misclassified as anomalous.

\subsubsection{Anomaly Scoring}
We define anomalies as motion deviations relative to the empirically established \textbf{normal} velocity range. Let \(\mathcal{C}_{\text{normal}}\) denote clusters assigned the \textbf{normal} semantic label. The boundary values are computed as:
\[
m_{\text{normal}}^{\min} = \min_{i \in \mathcal{C}_{\text{normal}}} {\min(C_i)} \quad \text{and} \quad
m_{\text{normal}}^{\max} = \max_{i \in \mathcal{C}_{\text{normal}}} {\max(C_i)}
\]

For a given motion magnitude \(m\), the anomaly score \(\mathcal{A}(m)\) is calculated as:
\[
\mathcal{A}(m) = 
\begin{cases}
\displaystyle\frac{m - m_{\text{normal}}^{\max}}{m_{\text{normal}}^{\max}} \times 100\%, & \text{if } m > m_{\text{normal}}^{\max} \\
\displaystyle\frac{m - m_{\text{normal}}^{\min}}{m_{\text{normal}}^{\min}} \times 100\%, & \text{if } m \leq m_{\text{normal}}^{\min} \\
0, & \text{otherwise}
\end{cases}
\]

This scoring mechanism assigns positive scores to unusually fast motion and negative scores to unusually slow motion, relative to what is considered normal for the crowd density. The approach provides intuitive, interpretable outputs while maintaining computational efficiency.

\section{Results and Analysis }
In this section, we first introduce our dataset collected from the Holy Mosque in Makkah, used to evaluate VelocityNet under realistic crowded conditions. We then present findings on optical flow performance, velocity modeling accuracy, and overall system efficiency.

%In this section, we first introduce our challenging dataset collected at the Holy Mosque in Makkah. We then present key findings from evaluating VelocityNet, covering optical flow module performance, velocity modeling methods, and overall system efficiency.

\begin{table}[!t]
    \centering
    \caption{Metadata tags per video (counts, average file size, average FPS)}
    \label{tab:tag_stats}
    \begin{tabular}{l c c c }
    \toprule
    \textbf{Tag}        & \textbf{Count} & \textbf{Avg Size (MB)} & \textbf{Avg FPS} \\ \midrule
    halt                & 10             & 20.92                   & 26.51            \\ 
    slow                & 9              & 21.99                   & 26.12            \\ 
    artifact            & 4              & 12.91                   & 29.96            \\ 
    group               & 5              & 18.19                   & 29.93            \\ 
    fast                & 6              & 16.77                   & 29.89            \\ 
    low-quality         & 9              & 15.58                   & 26.05            \\ 
    zoom-out            & 5              & 24.52                   & 25.98            \\ 
    running             & 5              & 12.35                   & 29.87            \\ 
    zoom-in             & 3              & 7.38                    & 24.81            \\ 
    lag (frame drop)                 & 1              & 27.67                   & 30.24            \\ \bottomrule
    \end{tabular}
\end{table}

\subsection{Dataset Overview}
We collected our dataset from video recordings at the Holy Mosque in Makkah, featuring exceptionally dense crowds with severe occlusions and highly constrained pedestrian motion—an inspiring real-world testbed for robust anomaly detection.

Crowd density is grouped into three levels:
\begin{enumerate}
  \item \textbf{High density:} Individuals cannot move freely due to severe congestion.
  \item \textbf{Medium density:} People move with some restriction; average distance between individuals is less than 2 m.
  \item \textbf{Low density:} Pedestrians move freely, maintaining average distances greater than 2m.
\end{enumerate}

% Motion analysis confirmed that walking speeds in low- and medium-density videos remain within typical ranges. In contrast, high-density recordings show a substantial reduction in pedestrian velocity due to crowd congestion and restricted movement. Our dataset comprises $15$ videos, averaging $18$ MB in size and $27.63$ FPS in frame rate.

Motion analysis confirmed that walking speeds in low- and medium-density videos remain within typical ranges. In contrast, high-density recordings show substantial reductions in pedestrian velocity due to crowd congestion and restricted movement. Our dataset comprises $15$ videos with normal and anomalous behaviors captured using varying camera setups, averaging $18$ MB in size and $27.63$ FPS.

Table \ref{tab:tag_stats} lists each video tag along with the count of videos, their average file size in megabytes, and average frame rate in frames per second.

\subsection{ Optical Flow Performance}

\begin{table*}[t]
    \centering    
    \scriptsize
    \resizebox{\textwidth}{!}
    {%
    \begin{tabular}{l *{9}{c}}
    \toprule
    \textbf{Model} &  \textbf{Params} &  \textbf{FLOPs} & \textbf{ Time(ms)-fp16} &  \textbf{Memory(GB)-fp16} &  \textbf{Time(ms)-fp32} &  \textbf{Memory(GB)-fp32} \\
    \midrule
     FlowFormer++ \cite{Shi_2023_CVPR} &  16.152 &  7257.856 &  497.254 &  6.554 &  905.917 &  12.882 \\
     VideoFlow\_mof \cite{Dong_2024_CVPR} &  13.453 &  7337.596 &   676.282 &  2.82 &  1075.577 &  5.343 \\
    RAPIDFlow \cite{morimitsu2024rapidflow} &  \textcolor{blue}{1.646} &  \textcolor{blue}{188.524} & \textcolor{blue}{40.610} &  \textcolor{blue}{0.492} & \textcolor{blue}{47.964} & \textcolor{blue}{0.724} \\
    RAFT \cite{teed2020raft} &  5.258 &  3357.219 &   142.031 & 1.443 & 239.582 & 2.503 \\
     Maskflownet \cite{Zhao_2020_CVPR} &  20.656 &  660.395 & 74.367 & 0.872 & 131.533 & 1.466 \\
     Skflow \cite{sun2022skflow} &  6.273 &  5933.491 &   477.687 & 1.833 & 753.861 & 4.054 \\
     Fastflownet \cite{kong2021fastflownet} &  \textbf{1.366} &  \textbf{49.698 }&  \textbf{30.839} & \textbf{0.421} & \textbf{40.323}& \textbf{0.523}\\
    \bottomrule
    \end{tabular}
    }
    \caption{Performance benchmark results for optical flow models evaluated on 1280×720 resolution images using an Nvidia A5000 GPU. All models were tested with 5 trials each, using FP32 or FP16 precision with warm-up enabled. Benchmarking conducted using PTLFlow \cite{morimitsu2021ptlflow} Framework.}
    \label{tab:resources}
\end{table*}

%\textbf{Choosing RAPIDFlow for Robust Real-World Performance:} \\
To select the optimal optical flow component for VelocityNet, we conducted comparative experiments evaluating several models across multiple performance metrics. Table \ref{tab:resources} compares optical flow models in terms of parameters, computational cost (FLOPs), inference speed, and memory consumption. While FastFlowNet exhibited the lowest runtime and resource usage, RAPIDFlow demonstrated superior accuracy and robustness in handling extremely dense crowd scenarios typical of our dataset.

Table \ref{tab:latency} further evaluates model latency and throughput. Although FastFlowNet achieved the lowest latency per frame, RAPIDFlow consistently delivered the highest throughput, ensuring robust real-time performance. Given these results, we selected RAPIDFlow as the optical flow backbone for VelocityNet due to its balanced accuracy and efficiency in dense crowd conditions.

%Table \ref{tab:resources} presents a comparative evaluation of Fastflownet model against existing baseline models in terms of parameter count, computational cost (FLOPs), inference time, and memory usage under both half-precision (fp16) and single-precision $fp_32$ settings. The results demonstrate that our model consistently achieves better runtime efficiency and lower memory consumption while maintaining a comparable or smaller parameter size and FLOPs. Notably, in both fp16 and fp32 configurations, our model exhibits faster inference times and reduced resource usage, indicating its suitability for deployment in resource-constrained environments.

\begin{table}[h]
    \centering    
    \begin{tabularx}{\linewidth}{@{}ll@{\hspace{3em}} c @{\hspace{1em}} c @{}}
    \toprule
    \textbf{Model} &  &\textbf{Latency } $\downarrow$ &  \textbf{Throughput } $\uparrow$ \\
    \midrule
     FlowFormer++ \cite{Shi_2023_CVPR} &  &0.360568 &  2.74 \\
     VideoFlow\_mof \cite{Dong_2024_CVPR} &  &N/A &  N/A  \\
     RAPIDFlow \cite{morimitsu2024rapidflow} &  &0.058226 &  \textbf{31.50}  \\ 
     RAFT \cite{teed2020raft}&  &0.147586 &  6.86 \\
     Maskflownet \cite{Zhao_2020_CVPR}&  &0.095802 &  10.67 \\
     Skflow \cite{sun2022skflow}&  &0.381284 &  2.56 \\
     Fastflownet \cite{kong2021fastflownet} &  &\textbf{0.040292} &  29.32 \\
    \bottomrule
    \end{tabularx}
    \caption{Optical flow model performance comparison showing latency and throughput metrics. Benchmarks conducted on Nvidia A5000 GPU with 5 trials per model. videoflow excluded due to out-of-memory errors persisting even at reduced 500×250 resolution.}
    \label{tab:latency}
\end{table}

% \subsection{Velocity Modeling Results}

% To determine the most effective method for predicting normal velocities from bounding box areas, we compared clustering approaches with linear and quadratic polynomial regression models. Figure \ref{fig:graph} illustrates these models alongside empirical data points. Our analysis showed that quadratic regression provided the most accurate representation of the nonlinear relationship between bounding box area and motion magnitude, especially at extreme scales. Thus, we adopted quadratic regression due to its superior performance for anomaly detection in densely crowded environments.

\subsection{Clustering Method Selection and Feature Relationship Analysis}
\begin{figure}[h]
  \centering
  \includegraphics[width=0.45\textwidth]{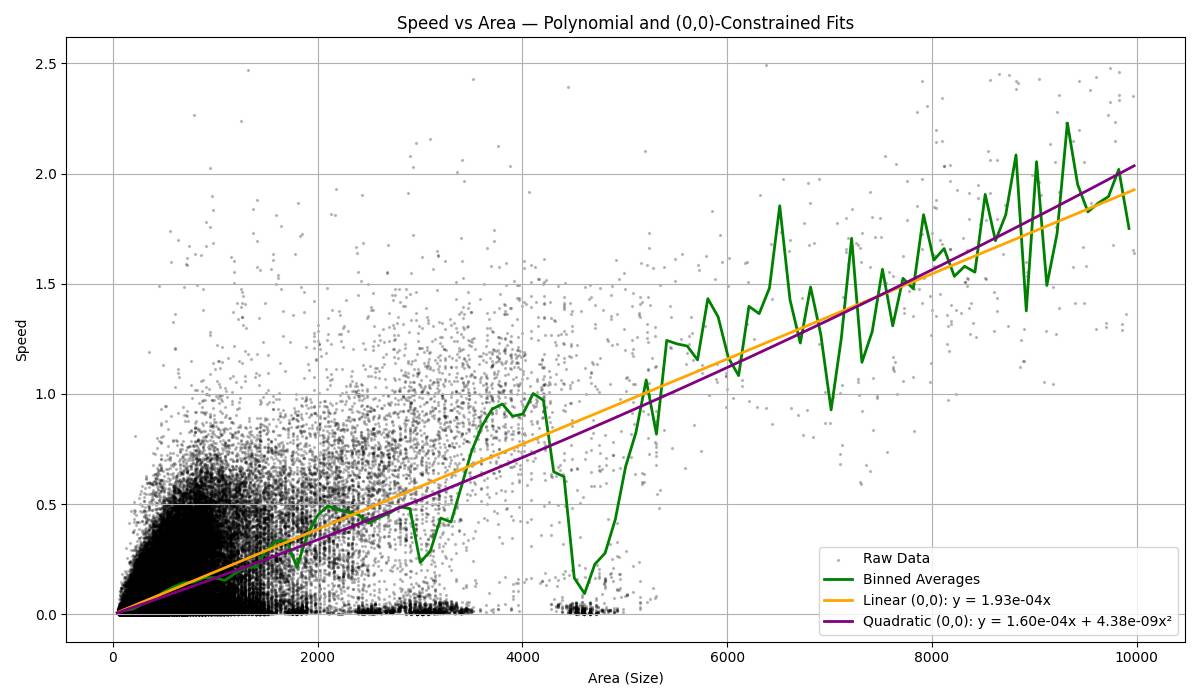}
  %\caption{An illustrative comparison of two regression models used to predict expected normal motion magnitude from bounding box area. The black dots represent individual data points, where the x-axis denotes bounding box area and the y-axis denotes measured motion magnitude. The green line shows the average magnitude in every area value. The yellow line corresponds to a linear polynomial fit, while the purple line represents a quadratic polynomial fit. Both models are constrained fits trained on the same dataset.}
  \caption{Comparison of linear (yellow) and quadratic (purple) regression models predicting motion magnitude from bounding box area (black dots).}
  \label{fig:graph}
\end{figure}

To determine the most effective method for predicting normal velocities from bounding box areas, we initially experimented with linear and quadratic polynomial regression models as potential clustering methods. Figure \ref{fig:graph} illustrates these regression models alongside empirical data points. While our analysis revealed that quadratic regression could accurately represent the nonlinear relationship between bounding box area and motion magnitude, particularly at extreme scales, the regression-based clustering approach proved inadequate due to its rigid assumptions of fixed functional relationships and limited flexibility in handling diverse crowd behavior patterns. The regression models could only create clusters based on curve residuals rather than utilizing the full multi-dimensional feature space effectively. Consequently, we adopted K-means clustering for our primary analysis due to its superior ability to identify natural groupings and robustness to data variability. However, the regression experiments provided valuable insights into the underlying relationships between velocity, bounding box size, and camera proximity, which informed our subsequent velocity preprocessing and anomaly detection methodology for densely crowded environments.

%Building upon the observed relationship between motion magnitude and bounding box area, we aimed to learn a predictive model that captures the expected "normal" velocity for a given object scale. Rather than applying a fixed normalization, we trained polynomial regression models on normal speed videos data to estimate the expected motion magnitude as a function of bounding box area.

%We experimented with both linear and quadratic polynomial models, where the input is the bounding box area $A$, and the output is the predicted normal motion magnitude $\tilde{m}$ . The quadratic model takes the form:

%\begin{equation}
%\tilde{m}(A) = \beta_0 + \beta_1 A + \beta_2 A^2
%\end{equation}

%and the linear baseline uses:
%\begin{equation}
%\tilde{m}(A) = \beta_0 + \beta_1 A
%\end{equation}

%Given a new observation with area $A_j$and measured motion magnitude $m_j$ we compute the deviation from the predicted value $\tilde{m}(A_j)$ and flag the instance as anomalous if the deviation exceeds a predefined threshold:

%\begin{equation}
%\left| m_j - \tilde{m}(A_j) \right| > \delta
%\end{equation}

%This threshold $\delta$ can be set empirically. Figure \ref{fig:graph} shows the training data alongside fitted curves for both linear and quadratic models. As illustrated, the quadratic model more accurately captures the curvature present in the data, particularly for extreme bounding box sizes.

\subsection{Visual Results and Runtime Performance}
%\textcolor{red}{Please update the title to be consistent. We are showing one thing in this section. I will prefere to use word like visual, Results, VelocityNet, our model etc }
Table \ref{tab:pipeline_results} presents the intermediate results of our proposed pipeline and its subprocesses. Given consecutive input frames $I_{t-1}$ and $I_t$, we first estimate the optical flow vectors using RAPIDFlow. Simultaneously, we apply head detection on the frame pairs to localize individuals within the scene. The extracted magnitudes for each detected person are then processed to compute velocity labels.

In our anomaly classification framework we define the thresholds for the four distinct behavioral categories based on hierarchical clustering results, for this experiment we set the following thresholds: \textbf{fast} anomalies ($m \geq 20$), \textbf{slow} anomalies ($-90 < m \leq -82$), \textbf{halt} behavior ($m \leq -90$), and \textbf{normal} behavior (all other cases). The final anomaly score is computed for each individual, and reporting decisions are made based on the above predefined anomaly reporting thresholds. The normal behavior is not highlighted within final output as we only highlight abnormal behaviors. 

While hierarchical clustering effectively groups velocities into semantic labels, we observe that cluster boundaries can exhibit marginal separation, particularly at label transitions. For example, the absolute maximum values of \textbf{halt} clusters and absolute minimum values of \textbf{slow} clusters demonstrate insufficient inter-cluster distance from their respective centroids. This proximity results in increased false positive rates when relying solely on hierarchical clustering for anomaly detection.

We addressed this limitation by leveraging the strengths of hierarchical clustering while mitigating its boundary sensitivity. Our method employs hierarchical clustering exclusively to identify \textbf{normal} behavior clusters, as empirical analysis demonstrates its robust capability to distinguish normal patterns from anomalous behaviors. Subsequently, we implement our anomaly scoring mechanism, which computes anomaly scores relative to the absolute boundaries of normal clusters (defined by the smallest minimum and largest maximum values).

This framework effectively automates the anomaly detection process while maintaining high precision in capturing normal behavioral patterns, which is the fundamental principle underlying anomaly detection. By establishing clear normal behavior baselines, our approach classifies any deviation as potentially anomalous. Additionally, this methodology prevents the reporting of trivial cases, such as velocity variations of merely $1\%$ below normal thresholds, thereby reducing false alarm rates in practical deployment scenarios.

% \begin{table*}
%   \centering
%   \footnotesize
%   \setlength{\tabcolsep}{3pt}
% \begin{tabular}{p{0.24\linewidth} p{0.24\linewidth} p{0.24\linewidth} p{0.24\linewidth} }
% \toprule
% \textbf{Input} & \textbf{Flow Map} & \textbf{Hierarchical Clustering} & \textbf{Output} \\
% \midrule 
% \includegraphics[width=\linewidth]{figures/inference_results/v4_raw.png} &
% \includegraphics[width=\linewidth]{figures/inference_results/v4_flow.jpg} &
% \includegraphics[width=\linewidth]{figures/inference_results/v4_hier.jpg} &
% \includegraphics[width=\linewidth]{figures/inference_results/v4_anom.jpg} \\
% \includegraphics[width=\linewidth]{figures/inference_results/v2_raw.png} &
% \includegraphics[width=\linewidth]{figures/inference_results/v2_flow.jpg} &
% \includegraphics[width=\linewidth]{figures/inference_results/v2_hier.jpg} &
% \includegraphics[width=\linewidth]{figures/inference_results/v2_anom.jpg} \\
% \bottomrule
% \end{tabular}

%   \caption{Qualitative results of VelocityNet (our model) on different crowd scenes and viewpoints. From left to right: input frame, generated flow map, hierarchical clustering model prediction, and anomaly pipeline final output.}
%   \label{tab:pipeline_results}
% \end{table*}

\begin{table*}
  \centering
  \footnotesize
  \setlength{\tabcolsep}{3pt}
\begin{tabular}{
  C{0.02\linewidth}  % Rotated label
  C{0.326\linewidth}  % Example 1
  C{0.326\linewidth}  % Example 2
  C{0.326\linewidth}  % Example 3
}
\toprule
\textbf{Stage} & \textbf{Outdoor Scene 1} & \textbf{Indoor Scene} &  \textbf{Outdoor Scene 2}\\
\midrule
\rotatebox{90}{Input} &
\includegraphics[width=0.95\linewidth]{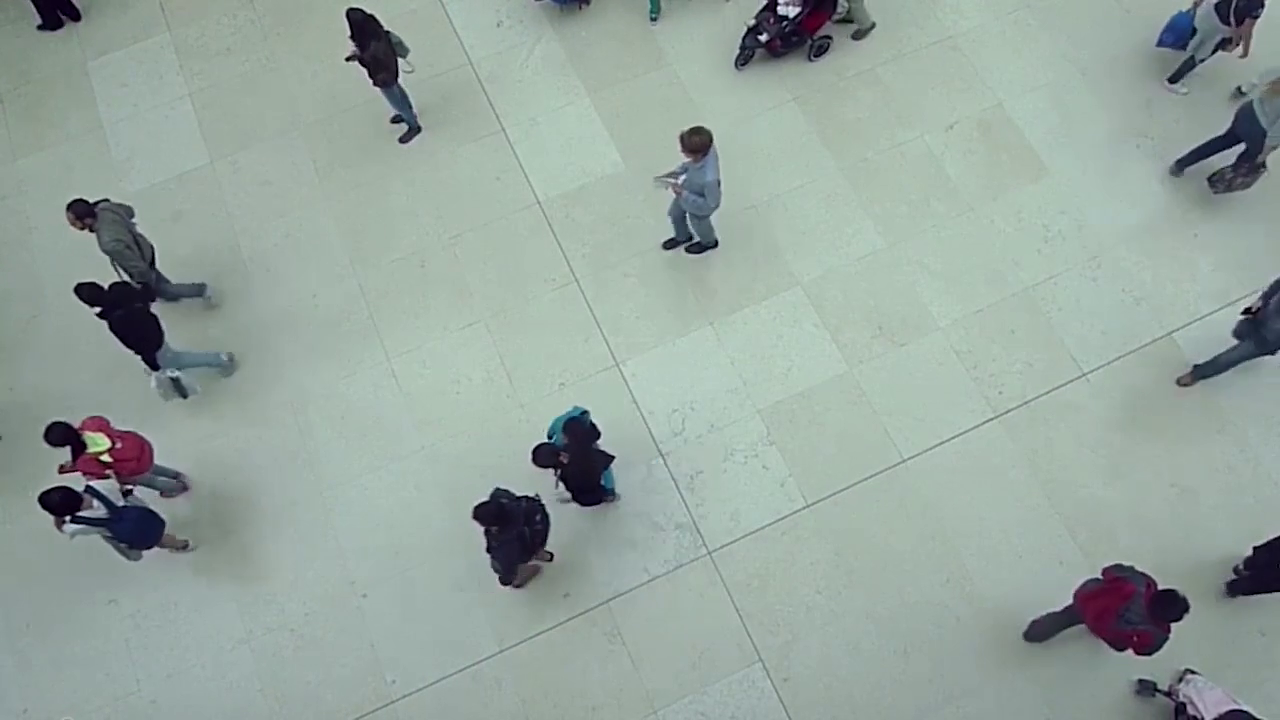} &
\includegraphics[width=0.95\linewidth]{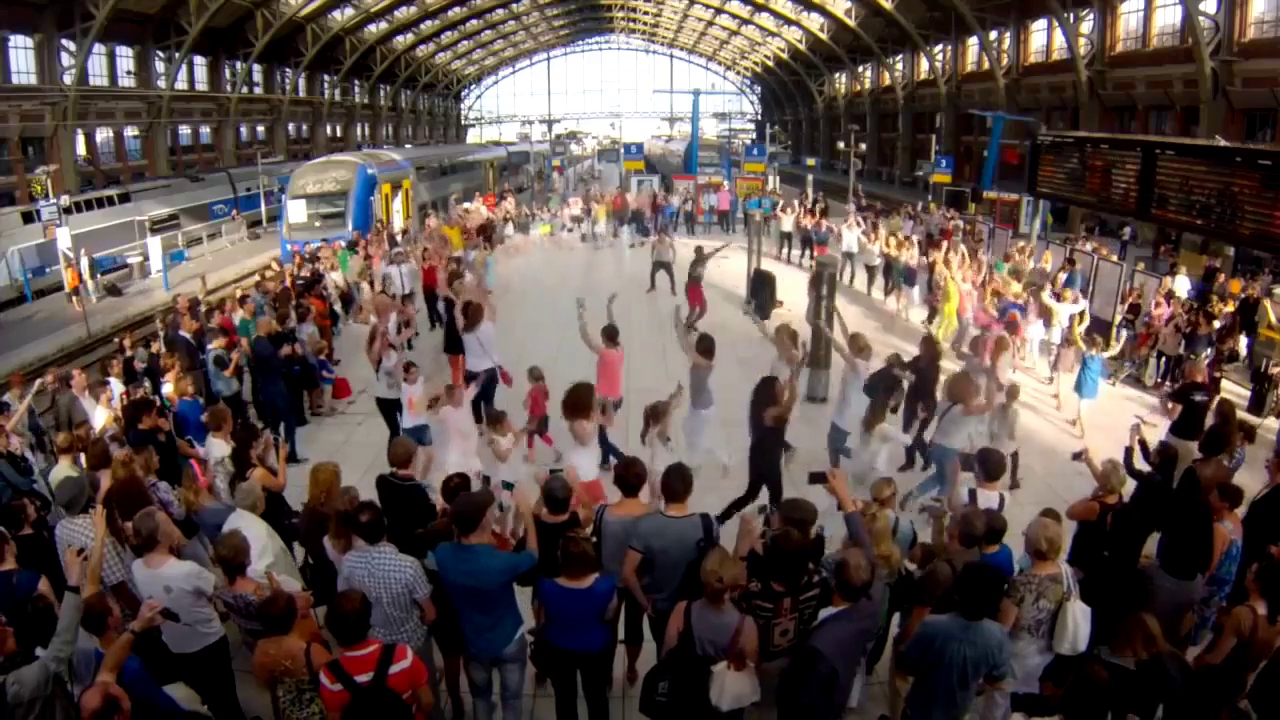} &
\includegraphics[width=0.95\linewidth]{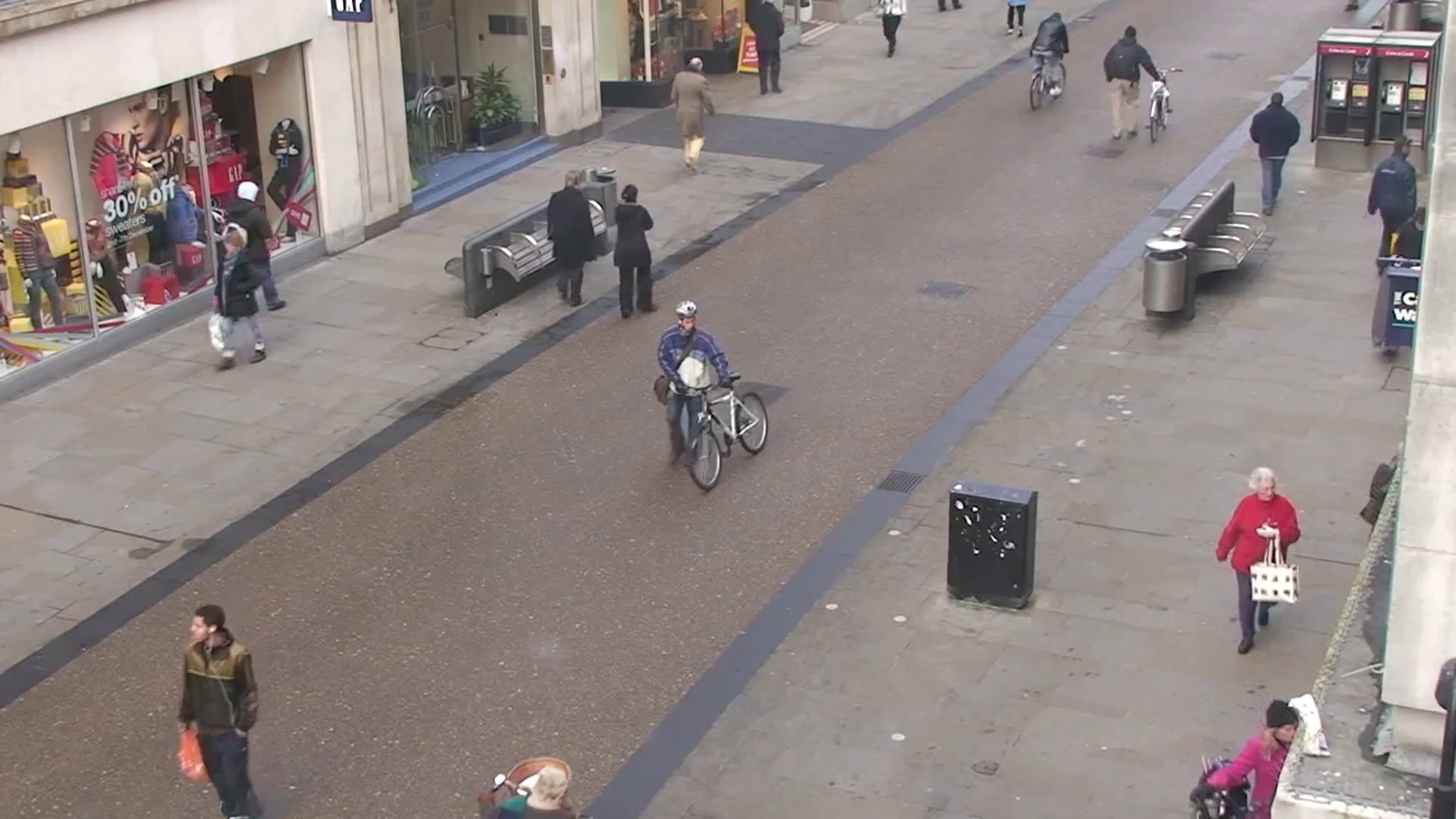} \\
\rotatebox{90}{Flow Map} &
\includegraphics[width=0.95\linewidth]{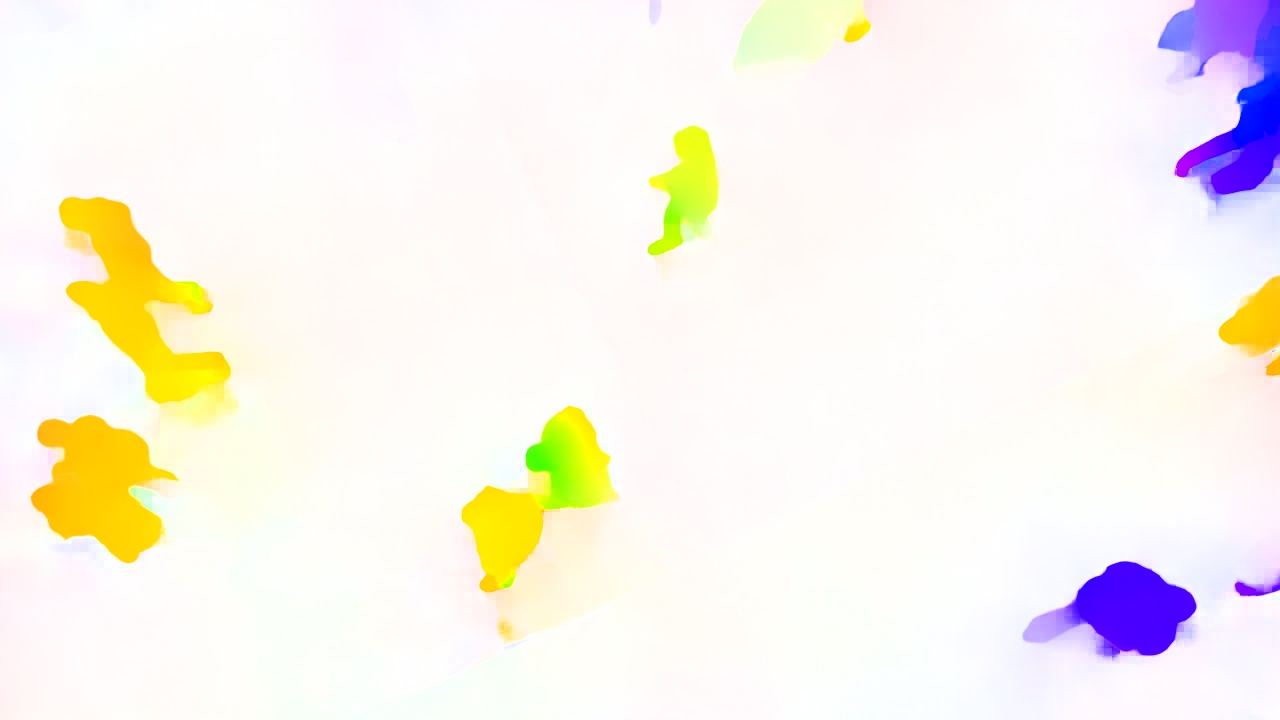} &
\includegraphics[width=0.95\linewidth]{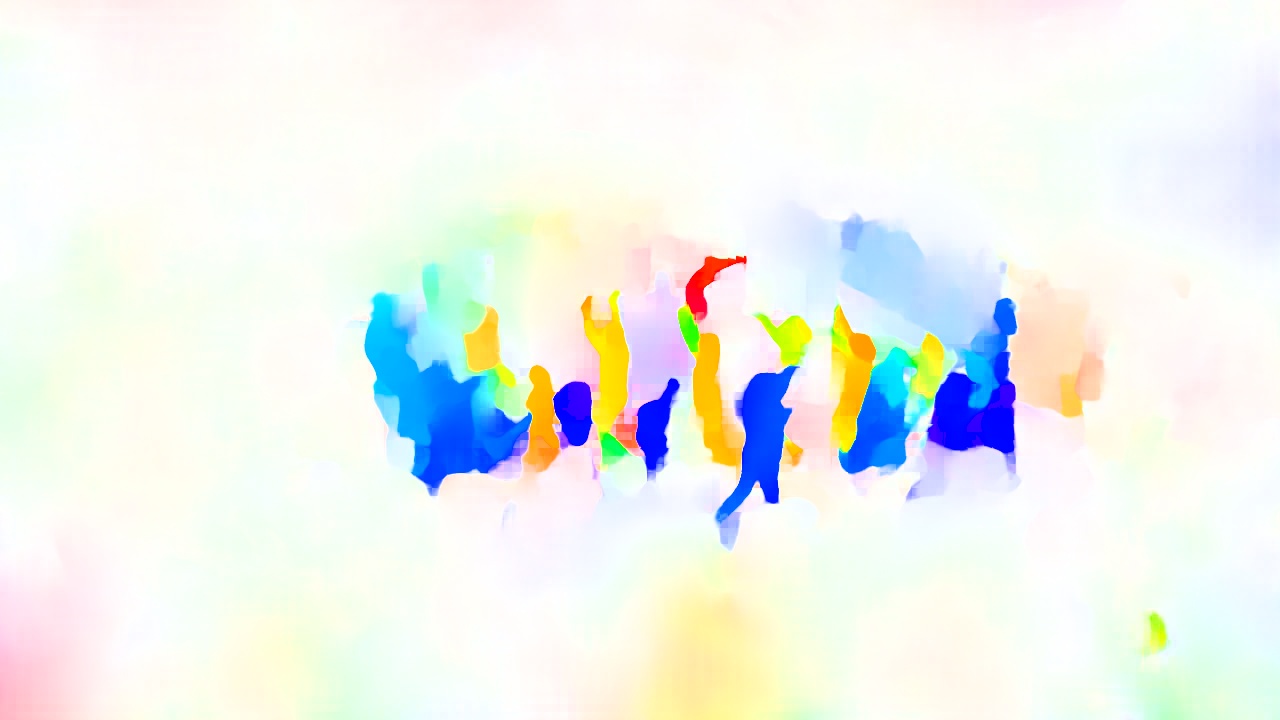}  &
\includegraphics[width=0.95\linewidth]{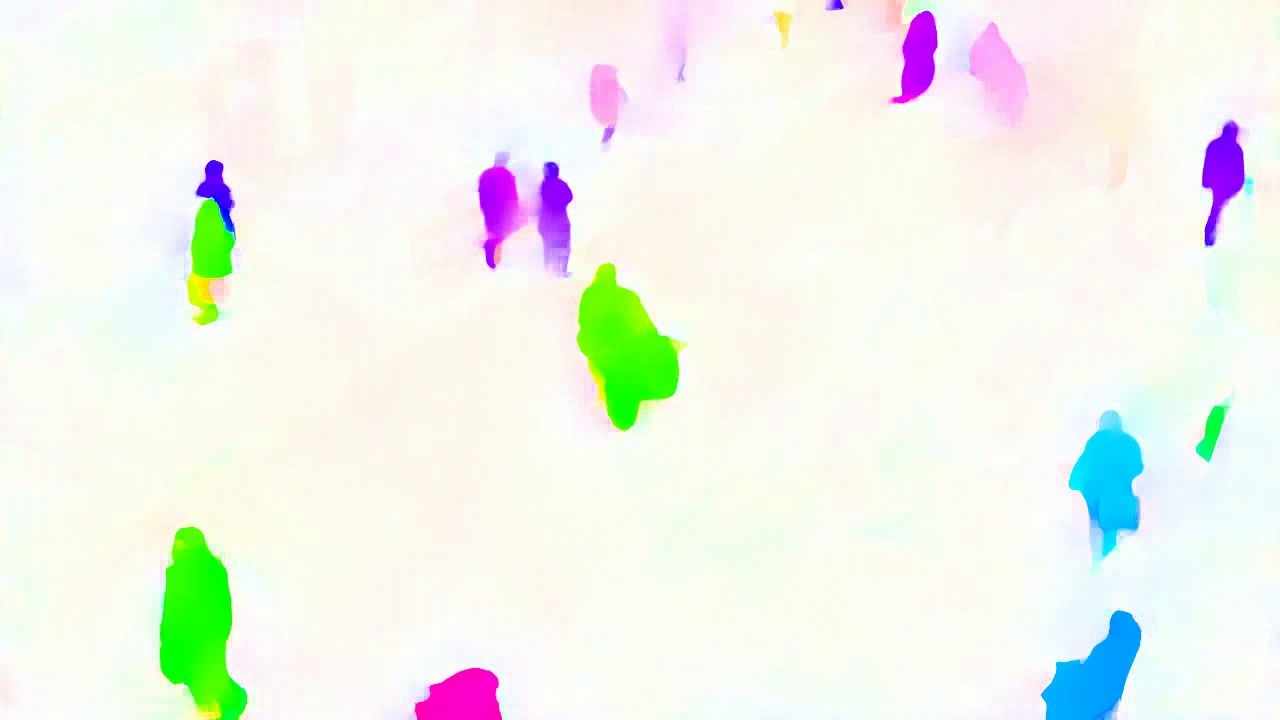} \\
\rotatebox{90}{Hierarchical Clustering} &
\includegraphics[width=0.95\linewidth]{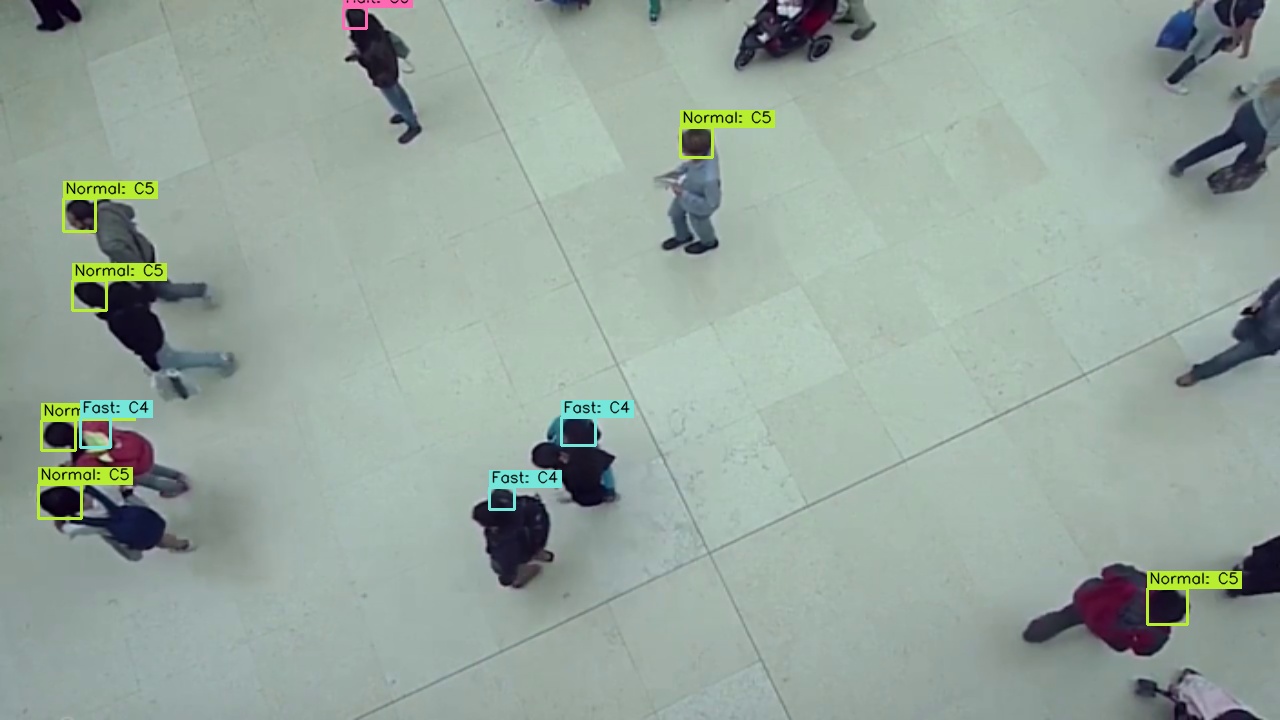} &
\includegraphics[width=0.95\linewidth]{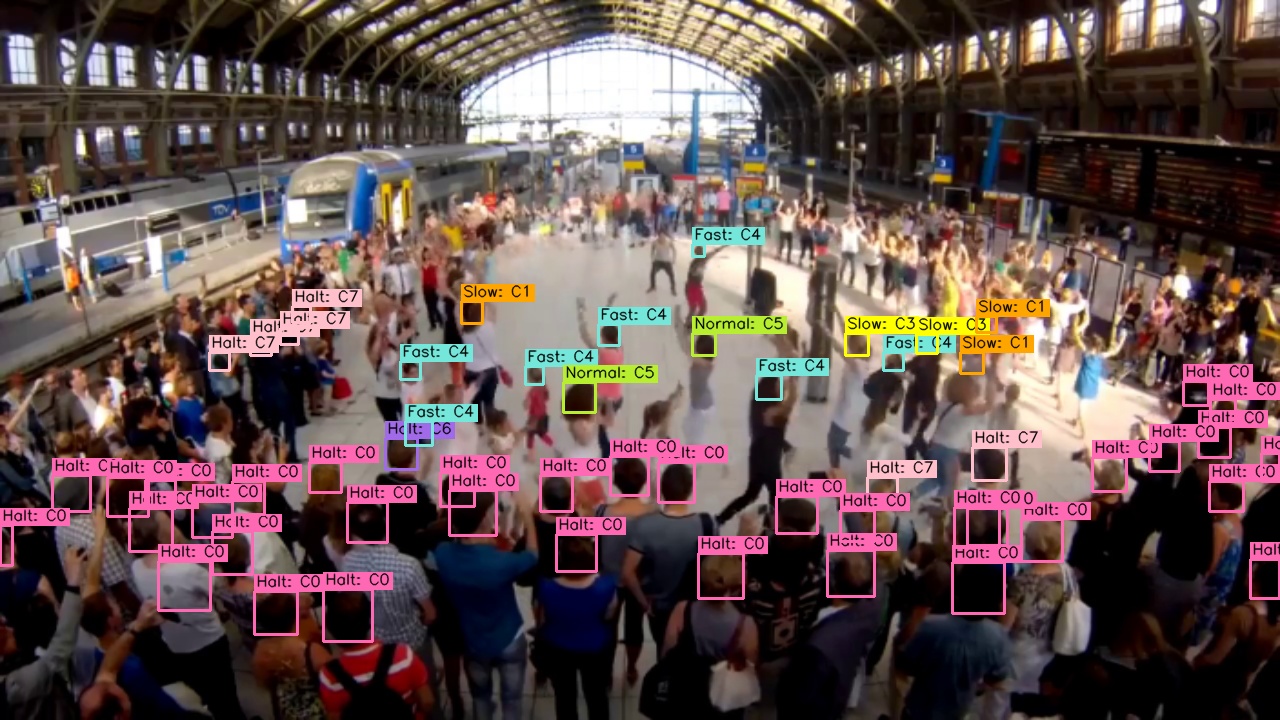}&
\includegraphics[width=0.95\linewidth]{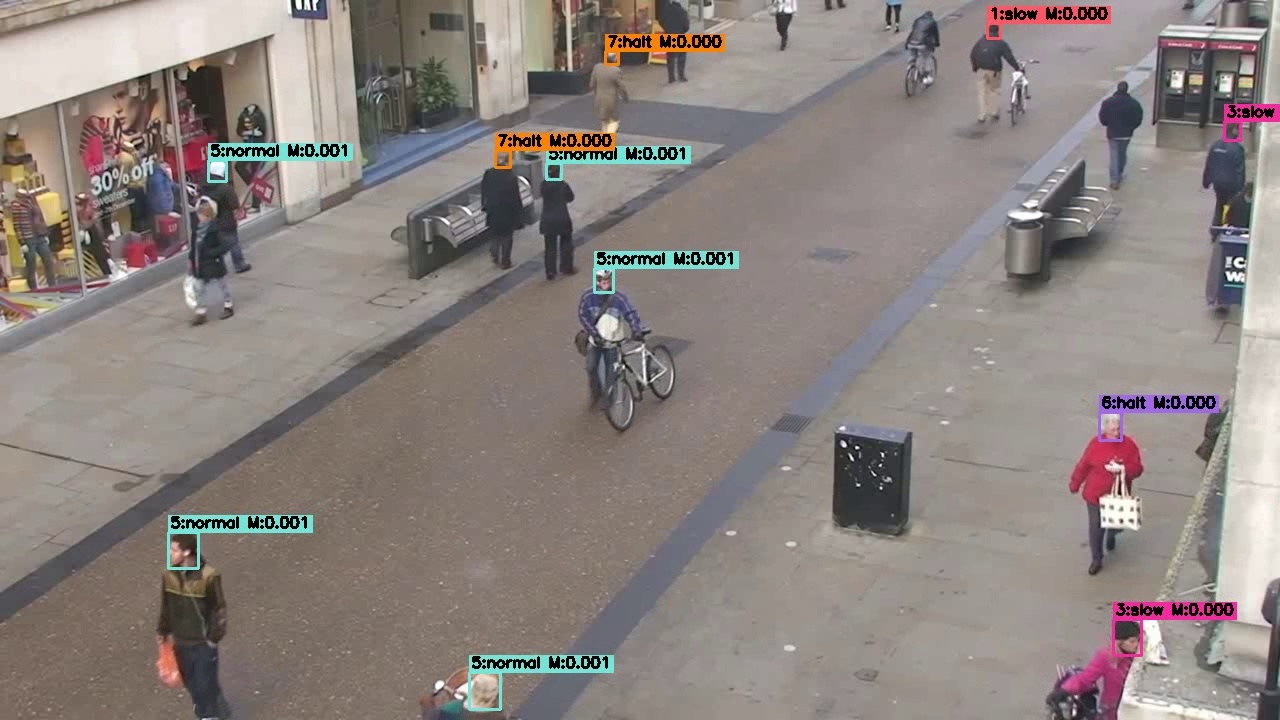} \\
\rotatebox{90}{Output} &
\includegraphics[width=0.95\linewidth]{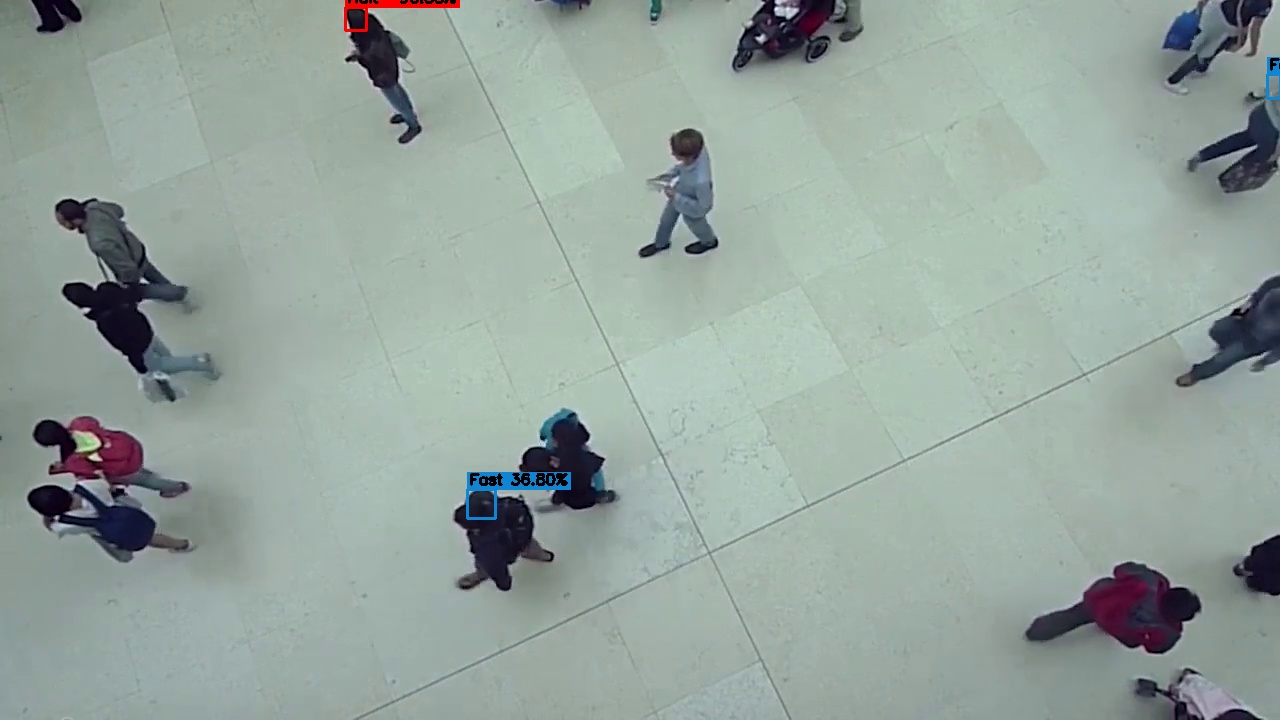} &
\includegraphics[width=0.95\linewidth]{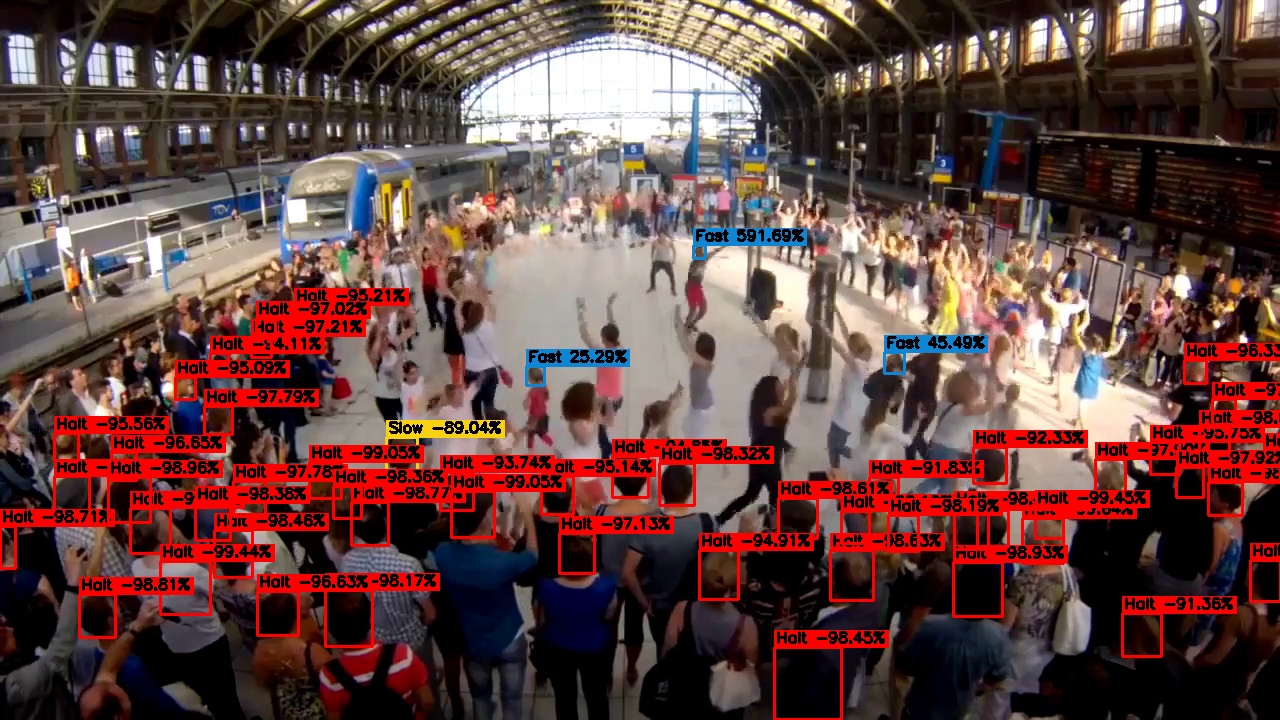}&
\includegraphics[width=0.95\linewidth]{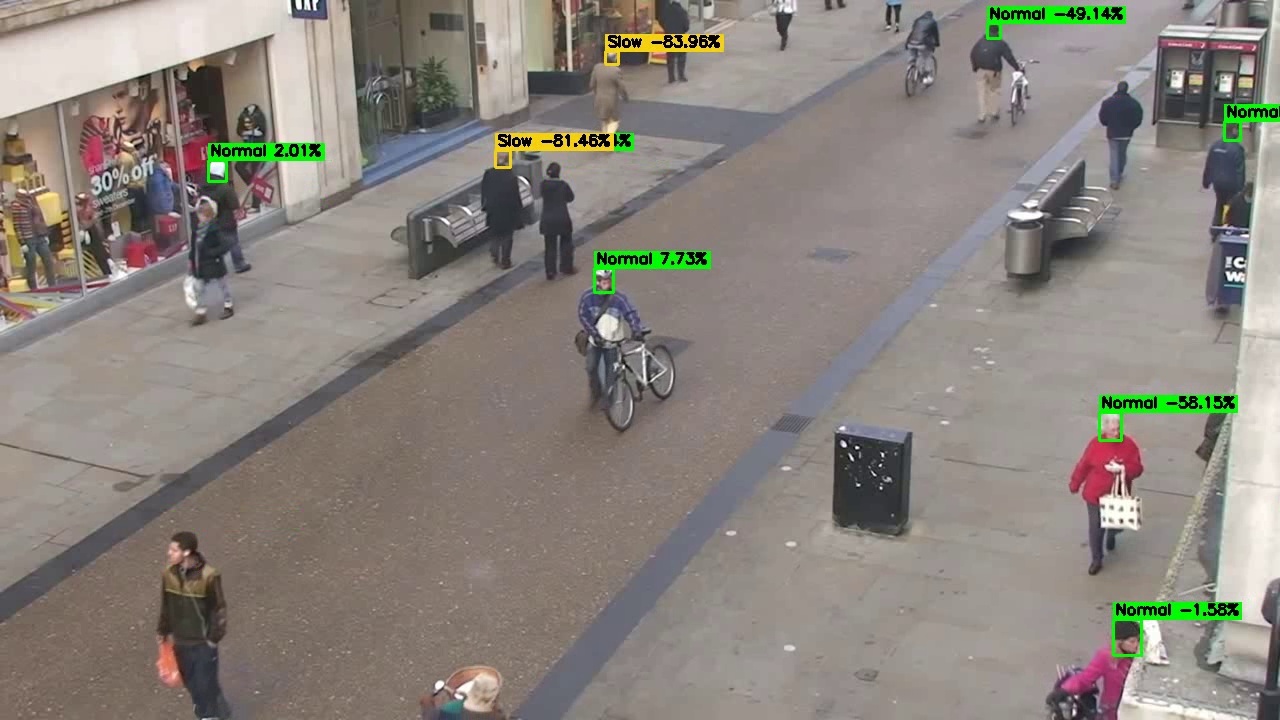}  \\
\bottomrule
\end{tabular}

\caption{Qualitative results of VelocityNet (our model) on different crowd scenes and viewpoints. From left to right: input frame, generated flow map, hierarchical clustering model prediction, and anomaly pipeline final output.}
\label{tab:pipeline_results}
\end{table*}

\label{sec:results}

\section{Conclusion}
\label{sec:conclusion}

In this paper, we introduced VelocityNet, a crowd anomaly detection framework utilizing a dual-pipeline approach combining head detection and dense optical flow. We employed hierarchical clustering to categorize velocities into semantic groups (halt, slow, normal, fast) and a percentile-based scoring mechanism to quantify deviations from typical motion patterns. This work serves as an initial step toward robust anomaly detection in densely crowded scenes. Future work will include evaluating VelocityNet on established anomaly detection benchmarks to test generalization capabilities, as well as benchmarking other state-of-the-art models on our challenging real-world dataset.

{
    \small
    \bibliographystyle{ieeenat_fullname}
    \bibliography{main}
}

\end{document}